\documentclass[lettersize,journal]{IEEEtran}
\usepackage{amsmath,amsfonts}
\usepackage{algorithmic}
\usepackage{algorithm}
\usepackage{algorithmic}
\usepackage{graphicx}
\usepackage{threeparttable}

\usepackage{amssymb}
\usepackage{makecell}
\usepackage{multirow}
\usepackage{booktabs}
\usepackage{color}
\usepackage{array}
\usepackage[caption=false,font=normalsize,labelfont=sf,textfont=sf]{subfig}
\usepackage{textcomp}
\usepackage{stfloats}
\usepackage{url}
\usepackage{verbatim}
\usepackage{cite}
\allowdisplaybreaks[4]
\newtheorem{remark}{Remark}

\usepackage{xr}
\begin{document}

\title{Knowledge-Assisted Dual-Stage Evolutionary Optimization of Large-Scale Crude Oil Scheduling}

\author{Wanting Zhang, Wei Du,~\IEEEmembership{Member,~IEEE,} Guo Yu,~\IEEEmembership{Member,~IEEE}, Renchu He, \\ Wenli Du, and Yaochu Jin,~\IEEEmembership{Fellow,~IEEE}
\thanks{Manuscript received 8 March 2023; revised 9 August 2023; accepted 18 September 2023.
This work was supported by National Natural Science Foundation of China (Key Program: 62136003), Major Program of Qingyuan Innovation Laboratory (Grant No. 00122002), National Natural Science Foundation of China (62173144, 62373154) and the Shanghai Committee of Science and Technology, China (Grant No.22DZ1101500). \textit{(Corresponding authors: Wei Du; Wenli Du.)}}
\thanks{Wanting Zhang, Wei Du, Renchu He, Wenli Du, and Yaochu Jin are with the Key Laboratory of Smart Manufacturing in Energy Chemical Process, Ministry of Education, East China University of Science and Technology, Shanghai 200237, China (e-mail: wt.zhang.ecust@gmail.com; duwei0203@gmail.com; renchuhe@ecust.edu.cn; wldu@ecust.edu.cn; yaochu.jin@uni-bielefeld.de).}
\thanks{Wenli Du is also with the Qingyuan Innovation Laboratory, Quanzhou 362801, China.}
\thanks{Yaochu Jin is also with the Chair of Nature Inspired Computing and Engineering, Faculty of Technology, Bielefeld University, 33619 Bielefeld, Germany.}
\thanks{Guo Yu is with the Institute of Intelligent Manufacturing, Nanjing Tech University, Nanjing 211816, China (e-mail: gysearch@163.com).}}

\markboth{IEEE TRANSACTIONS ON EMERGING TOPICS IN COMPUTATIONAL INTELLIGENCE}
{}

\maketitle

\begin{abstract}
With the scaling up of crude oil scheduling in modern refineries, large-scale crude oil scheduling problems (LSCOSPs) emerge with thousands of binary variables and non-linear constraints, which are challenging to be optimized by traditional optimization methods. To solve LSCOSPs, we take the practical crude oil scheduling from a marine-access refinery as an example and start with modeling LSCOSPs from crude unloading, transportation, crude distillation unit processing, and inventory management of intermediate products. On the basis of the proposed model, a dual-stage evolutionary algorithm driven by heuristic rules (denoted by DSEA/HR) is developed, where the dual-stage search mechanism consists of global search and local refinement. In the global search stage, we devise several heuristic rules based on the empirical operating knowledge to generate a well-performing initial population and accelerate convergence in the mixed variables space. In the local refinement stage, a repair strategy is proposed to move the infeasible solutions towards feasible regions by further optimizing the local continuous variables. During the whole evolutionary process, the proposed dual-stage framework plays a crucial role in balancing exploration and exploitation. Experimental results have shown that DSEA/HR outperforms the state-of-the-art and widely-used mathematical programming methods and metaheuristic algorithms on LSCOSP instances within a reasonable time.
\end{abstract}

\begin{IEEEkeywords}
Large-scale crude oil scheduling, evolutionary optimization, heuristic rules, local refinement
\end{IEEEkeywords}

\section{Introduction}\label{section 1}
\IEEEPARstart{P}{roduction} scheduling plays a vital role in improving the economic performance and competitiveness of refineries \cite{Shah2011}. Short-term crude oil scheduling is a leading segment for the overall refinery scheduling. It aims to make the short-term decision for a week or ten days, depending on the operation status and material transfer. The scheduling quality directly affects the stability of subsequent production. However, scheduling optimization for crude oil is challenging with the interaction of discrete events and continuous processes. Currently, crude scheduling still relies on the experience of schedulers due to the lack of effective techniques or software tools \cite{Li2020}. Therefore, short-term scheduling optimization for crude oil has attracted extensive attention from both academics and industries.

Typically, mathematical programming (MP) models represented with discrete- \cite{Saharidis2009} or continuous-time \cite{Zhao2016} are widely formulated to describe short-term scheduling problems of crude oil. The discrete models divide the whole scheduling horizon into several periods with uniform time slots. The operating activities that begin or end at the slots are represented by binary variables. It is convenient for complex scheduling problems to model as discrete-time representations. However, the commonly-used way to improve the accuracy by using the small slots may lead to a considerable increase of binary variables, which brings difficulties in computing. The continuous-time models using fewer binary variables in single \cite{Reddy2004b} or multiple time-grids models \cite{cerda2015,Zhao2016} have been reported to overcome this deficiency. Nevertheless, the number of events for operations in these models is required to be given in advance and only determined by trial and error, which is difficult to implement in real-world applications. As a result, the practicality and optimality of the modeling method are suggested to be balanced in large-scale complex problems.

In addition, the blending constraints involving bilinear terms in crude oil scheduling optimization lead to mixed-integer nonlinear programming (MINLP). As discussed in \cite{Li2007}, current commercial solvers may fail to converge to an optimal solution when solving MINLP problems. In order to solve this problem efficiently, several strategies have been reported. The common way is to solve mixed-integer linear programming (MILP) and nonlinear programming (NLP) iteratively \cite{Mouret2011, Assis2019, Assis2021}. However, this approach may not find a feasible solution due to the composition concentration discrepancy caused by the relaxation of nonlinear terms \cite{Yadav2012}. Similarly, the work in \cite{Reddy2004a} presents a rolling-horizon framework, which eliminates the composition discrepancy, but it may fail to get a feasible schedule since myopic behavior \cite{cerda2015}. Therefore, despite the fact that MP methods have been used extensively, it compromises the originality of complex problems and the quality of solutions.

As introduced above, most research based on MP methods pays attention to attaining a better schedule with fewer binary variables, nonlinear terms, and less CPU time. However, with the increase in scheduling scale, the number of discrete variables and nonlinear terms expands significantly, which leads to an exponential growth of computational time \cite{Panda2018}. In recent years, metaheuristic algorithms as a powerful tool have been introduced to solve nonlinear \cite{Das2011} and large-scale \cite{Tian2021} optimization problems. These advantages underscore the superiority of metaheuristics over traditional methods in various production scheduling problems.
Numerous studies have examined the use of metaheuristics in scheduling, as reviewed in references \cite{Li2022,Gao2019,Branke2016}. More specifically, Pan \textit{et al.} \cite{Pan2022} explored flexible job-shop scheduling and developed a bi-population evolutionary algorithm with a feedback mechanism for energy-efficient optimization. To address the exponential complexity in distributed flow-shop scheduling optimization, Wang \textit{et al.} \cite{Wang2021} proposed a metaheuristic algorithm based on a knowledge-based cooperative strategy, which effectively tackles this strongly NP-hard problem. Furthermore, Du \textit{et al.} \cite{Du2019} introduced a decision variable classification to support evolutionary optimization in solving high-dimensional order scheduling problems. Evidently, these studies demonstrate that metaheuristics have emerged as effective approaches for handling complex scheduling problems.
In addition, another attractive feature of metaheuristics compared to MP is that they can be easily incorporated with sophisticated simulation models that introduce many realistic details. Specifically, these details contain constraints that cannot clearly be modeled by equations or would give rise to models with a large number of variables but relatively few free decision variables.

Motivated by the advantages of metaheuristics, a number of evolutionary algorithms (EAs) have been investigated on crude oil scheduling problems. For instance, a structural adaptive genetic algorithm (SAGA) for crude scheduling problems is proposed in \cite{Ramteke2012} and further applied to the optimization for the robust scheduling of crude oil \cite{Panda2018, Panda2019}. Though this graph-based representation used in this algorithm provides a sparse problem representation, the convergence of SAGA remains slower in large-scale problems with a huge number of decision variables. The work in \cite{Hou2017,HOU2020} investigates the multi-objective optimization problem for crude oil scheduling in an inland refinery and applies GA to solve the assignment of charging tanks and distillers. However, their methods are only implemented on condition that storage tanks have sufficient inventory. The realistic scale scheduling problems are far from being resolved. While crude oil scheduling problems have attracted extensive EA-based research, there is a notable lack of emphasis on the utilization of knowledge. In general, specific knowledge can be derived from the characteristics of problems, and its application can significantly enhance search efficiency for real-world problems \cite{Jin2005}. In the context of crude oil scheduling problems, empirical knowledge embedded within operations, such as the preference for selecting crude types during the blending process, is easily understandable and extractable. Consequently, more efforts should be made to tackle large-scale complex scheduling based on metaheuristics and improve the applicability of approaches for practical operations.

Based on the above discussions, it can be found that there is still a big gap between the existing theoretical research and actual processes in large-scale crude scheduling optimization. To this end, in this paper, we propose a novel dual-stage evolutionary algorithm driven by heuristic rules (DSEA/HR) to efficiently address the large-scale crude oil scheduling problems (LSCOSPs) in the real-world refinery. The contributions of this work are listed as follows:

\begin{enumerate}
	\item In contrast to the small or medium-scale problems resolved in existing studies, large-scale crude oil scheduling problems are modeled from crude unloading, transportation, crude distillation unit processing, inventory management of intermediate products to the practical processing and operating constraints.

	\item In the DSEA/HR, the dual-stage search consists of global search and local refinement. The former is to speed up the optimization process, while the latter is to reform the infeasible solutions. Particularly, we extract problem-specific knowledge by analyzing the impact of crude blending operations on search space. This knowledge is utilized to formulate two heuristic rules for population initialization, thereby improving the global search efficiency within the mixed decision space. \label{R1.1.2}

	\item The proposed DSEA/HR has been successfully implemented on practical LSCOSP instances, and the experimental results have shown that our approach is superior to the compared methods in performance and computational efficiency.
\end{enumerate}

The remainder of the paper is organized as follows. Section \ref{section 2} formulates the mathematical model of the LSCOSP.  The proposed DSEA/HR approach based on heuristic rules and a repair strategy is presented in Section \ref{section 3}. Section \ref{section 4} reports the experimental results on real-world cases, together with performance analysis. Finally, the paper is concluded in Section \ref{section 5}.

\section{Large-Scale Crude Oil Scheduling Problem} \label{section 2}
\subsection{Problem Description}
\begin{figure}[!t]
	\centering
	\includegraphics[scale= 0.265]{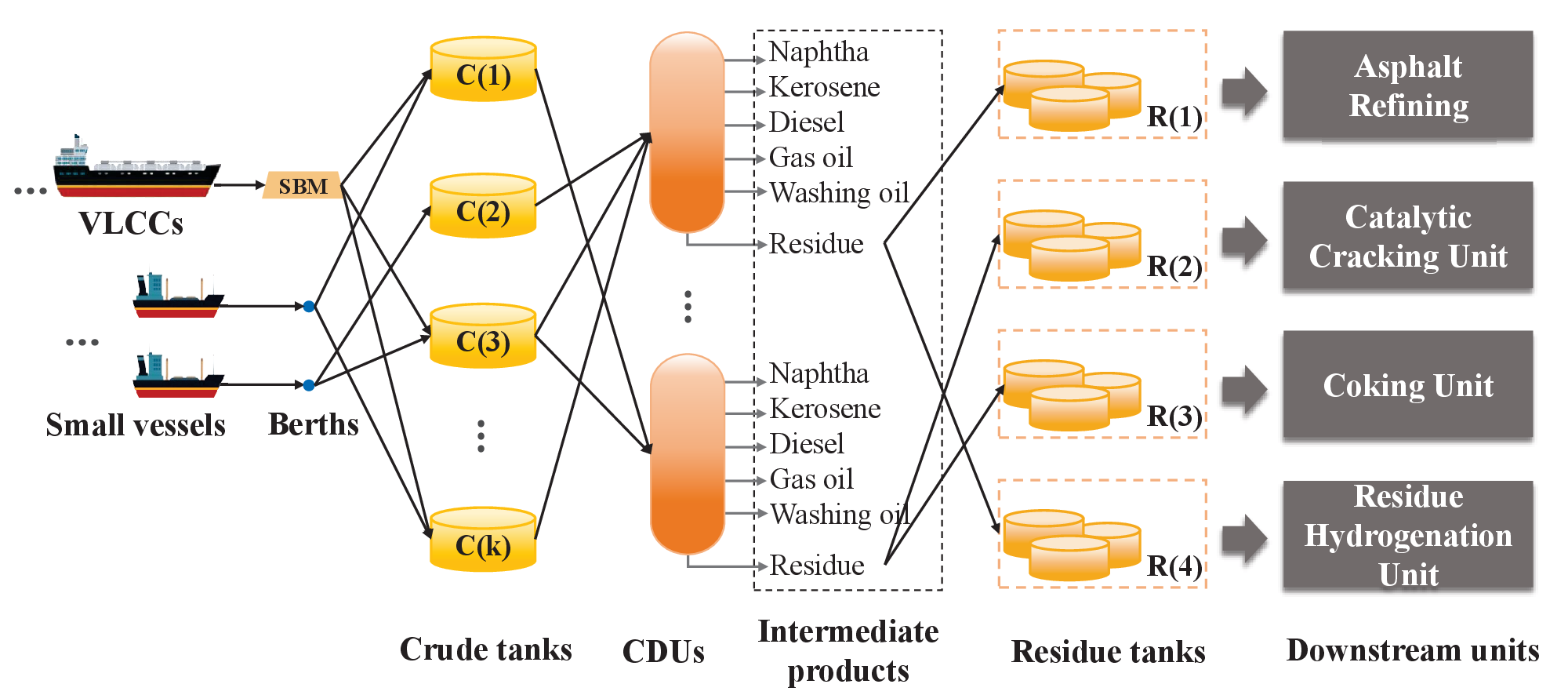}
	\caption{Crude oil scheduling structure of marine-access refineries.}
	\label{structure}
\end{figure}

The scope of the studied LSCOSP is illustrated in Fig. \ref{structure}. Firstly, crude oil is transferred to tanks from very large crude carriers (VLCCs) or small-size vessels. Then the crude from different tanks is blended and charged to crude distillation units (CDUs) where various intermediate products such as residue, diesel, and naphtha are produced. Finally, these intermediates are fed to specific downstream units. Notably, we focus on a marine-access refinery in this paper. The range of component concentration in the tank fluctuates widely due to marine-access refineries without dedicated terminal storage tanks. Compared with the scheduling of inland refineries, severe composition discrepancy will arise due to the relaxation of nonlinear terms in deterministic methods. In addition, the residue may exceed loads of secondary units by overusing inexpensive heavy crudes \cite{Dai2021}. To address this issue, we have also taken into account the inventory of four types of residue tanks. The residue types are related to the operating modes in CDUs and crude types. Apparently, the above operational features undoubtedly make resource allocation more complex.

\subsection{Problem Formulation} \label{formulation}
Based on the above problem description, we present the mathematical model of the LSCOSP in this section, utilizing the notations listed in Table \ref{notations}. It is worth noting that we chose the discrete-time representation of the LSCOSP for the following reasons: (1) this representation is extensive and effective for complex industrial problems on a large scale \cite{Wassick2011}; (2) the discrete-time representation conforms with the conditions of the marine-access refineries where the practical operations only occur at regular time points. Next, we formulate the constraints and objective function as follows.

\begin{table}[!t]
	\centering
	\small
	\caption{Nomenclature.}\label{notations}
	\resizebox{\linewidth}{!}{
		\begin{tabular}{p{2.2cm}p{8.5cm}}
			\hline
			Notation & Meaning \\
			\hline
			\textbf{\textit{Indices and sets:}} & \\
			$v$ & Index of vessels, $v \in \{1,...,V\}$ \\
			$t$ & Index of tanks, $t \in \{1,...,T\}$ \\
			$c$ & Index of crude types, $c \in \{1,...,C\}$ \\
			$k$ & Index of crude properties, $k \in \{1,...,K\}$ \\
			$u$ & Index of CDUs, $u \in \{1,...,U\}$ \\
			$r$ & Index of residue types, $r \in \{1,...,R\}$ \\
			$s$ &Index of product types, $s \in \{1,...,S\}$ \\
			$n$ & Index of periods, $t \in \{1,...,N\}$ \\
			$RC_r$ &Set of crude allowed to product residue $r$ \\
			\textbf{\textit{Parameters:}} & \\
			$AD_v$ & Arrival day of vessel $v$ \\
			${MT}_u$ &  {The maximum number of the charging tank connected} by CDU $u$ in each period\\
			${CR}_{r,n}$ & {Consumption rate of residue type $r$ in period $n$} \\
			${FU}_u^L$ &  {Lower bound of crude received by CDU $u$ in one period} \\
			${FU}_u^U$ &  {Upper bound of crude received by CDU $u$ in one period} \\
			$P_{c,k}$ & Key property $k$ of crude $c$ \\
			$P_{u,k}^L$ & Lower bound of feed property $k$ of CDU $u$ \\
			$P_{u,k}^U$ & Upper bound of feed property $k$ of CDU $u$ \\
			${TL}_{t}^L$ & Lower bound of liquid level of crude tank $t$ \\
			${TL}_{t}^U$ & Upper bound of liquid level of crude tank $t$ \\
			${IR}_{r}^L$ & Lower bound of inventory of residue $r$ \\
			${IR}_{r}^U$ & Upper bound of inventory of residue $r$ \\
			${TS}_t$ & Area of bottom of tank $t$ \\
			${yield}_{u,c,s}$ & Product $s$ yield of crude $c$ in CDU $u$\\
			${FO}_{s,u}^L$ &  {Lower bound of the outflow of product $s$ from CDU $u$}\\
			${FO}_{s,u}^U$ &  {Upper bound of the outflow of product $s$ from CDU $u$}\\
			\multicolumn{2}{l}{\textbf{\textit{Binary variables:}}} \\
			${XC}_{v,c,t,n}$ & =1,  {if vessel $v$ transfers crude $c$ to tank $t$ in period $n$,} 0 otherwise \\
			${XT}_{t,u,n}$ & =1, {if tank $t$ is charging to CDU $u$ in period $n$,} 0 otherwise \\
			${XU}_{u,r,n}$ & =1, if CDU $u$ produces residue type $r$ in period $n$, 0 otherwise \\
			\multicolumn{2}{l}{\textbf{\textit{Continuous variables:}}} \\
			${CTU}_{t,u,n}$  &  {0-1 continuous variables; =1, if connection status of} {tank $t$ to CDU $u$ changes between period $n-1$ and period $n$,} 0 otherwise \\
			${CUR}_{u,r,n}$  &  {0-1 continuous variables; =1, if connection status of } {residue type $r$ in CDU $u$ changes between period $n-1$}  and period $n$, 0 otherwise\\
			$CO_{u,n}$ & {0-1 continuous variables; =1, if connection status } changes between period $n\!-\!1$ and period $n$, 0 otherwise \\
			${IV}_{v,c,n}$ &  {Amount of crude $c$ in vessel $v$ at the end of period $n$ }\\
			${ITC}_{t,c,n}$ &  {Amount of crude $c$ in tank $t$ at the end of period $n$} \\
			${FV}_{v,c,t,n}$ &  {Amount of crude $c$ unloaded by vessel $v$ to tank $t$ at the}  end of period $n$ \\
			${FR}_{u,r,n}$ &  {Weight of reside type $\!r\!$ produced by CDU $\!u\!$ at the end}  of period $n$ \\
			${IT}_{t,n}$ & Inventory of crude tank $t$ at the end of period $n$ \\
			${IR}_{r,n}$ & Inventory of residue $r$ at the end of period $n$ \\
			${FTC}_{t,c,u,n}$ &  {Amount of crude $c$ charged by tank $t$ to CDU $u$ in} period $n$\\
			${FUC}_{u,c,n}$ & Total amount of crude $c$ received by CDU $u$ in period $n$ \\
			${FU}_{u,n}$ & Total feed of CDU $u$ in period $n$\\
			${FT}_{t,u,n}$ & Total amount of crude charged by tank $t$ to CDU $u$  in period $n$ \\
			$FTC_{t,c,u,n}$ & Amount of crude $c$ received by CDU $u$ from tank $t$  in period $n$ \\
			${FO}_{u,s,n}$ & Weight of product $s$ output from CDU $u$ in period $n$\\
			\hline
		\end{tabular}
	}
\end{table}

\subsubsection{Constraints}
\ \newline \indent
In the LSCOSP, two groups of constraints based on discrete events and continuous processes are concluded:
\begin{itemize}
\item Operating constraints, which are denoted by the binary variables representing the operating requirements.
\item Transfer constraints, which include material balance, capacity, and processing conditions.
\end{itemize}

Given that a complete mathematical model involves numerous formulations, in order to improve the readability of this paper, we have provided the primary constraints, while the remaining constraints can be found in Section S.I of the supplementary file.

\begin{itemize}
	\item [a)] {\textit{Operating constraints: }}
	\begin{align}
		&\scalebox{0.9}{$\sum\limits_{v \in V}\sum\limits_{c \in C} {XC}_{v,c,t,n} \le 1,\forall t,n,$}	\label{a-1}\\
		&\scalebox{0.9}{${XC}_{v,c,t,n} + {XT}_{t,u,n} \le 1,\forall v,c,t,u,n,$}	\label{a-2} \\
		&\scalebox{0.9}{$\sum\limits_{u \in U} {XT}_{t,u,n} \le 2,\forall t,n,$}	\label{a-3} \\
		&\scalebox{0.9}{$\sum\limits_{t \in T} {XT}_{t,u,n} \le {MT}_u,\forall u,n,$}  \label{a-4} \\
		&\scalebox{0.9}{$\sum\limits_{r \in R} {XU}_{u,r,n} \le 1,\forall u,n,$} \label{a-5}
	\end{align}
where Eq. \eqref{a-1} specifies that each crude tank is capable of receiving only one type of crude from a vessel during each period. This constraint is in place to prevent the mixing of different quality crudes in the same tank during unloading. Eq. \eqref{a-2} ensures that the inlet and outlet operations of each tank cannot be implemented simultaneously. Eq. \eqref{a-3} means that no more than two CDUs can charge a single tank simultaneously. Eq. \eqref{a-4} defines the maximum number of the charging tank that can be used. Eq. \eqref{a-5} indicates that each CDU can only produce one type of residue during each period. The remaining operating constraints are provided in Eqs. (S.1)-(S.11) of the supplementary file.
\item [b)] {\textit{Transfer constraints: }}
\begin{align}
	&\scalebox{0.9}{${FT}_{\!t,u,n} \!\cdot\! ITC_{t,c,n\!-\!1}\! = \! FTC_{t,c,u,n} \!\cdot\! IT_{t,n\!-\!1}, \!\forall t,u,c,n,$} \label{b-1} \\
	&\scalebox{0.9}{$P_{\!u,k}^L  \!\!\le \!\!	\sum\limits_{c \in C} \!\left( {FUC}_{u,c,n}  \!\cdot\! {P_{\!c,k}}\right)\! /\!{\sum\limits_{c \in C} \!{FUC}_{u,c,n} } \!\!\le\!\! P_{\!u,k}^U, \forall u,k,n,$} \label{b-2}
\end{align}

where Eq. \eqref{b-1} assures that the concentration of the crude mix composition in the tanks is equal to that in the CDU feed streams. Noteworthy, this equation contains two non-convex bilinear terms, leading to poor convergence in large-scale problems. Eq. \eqref{b-2} requires that the key properties (i.e., density, sulfur content, and total acid number) of the feed should be satisfied in the allowable range after the crude blending process. The remaining transfer constraints are listed in Eqs. (S.12)-(S.23) of the supplementary file.
\end{itemize}
\subsubsection{Objective function} \label{obj}
Minimizing the total operating costs is commonly adopted as the optimization objective, which includes the vessel waiting cost, the inventory cost, and the changeover cost \cite{Yadav2012}. To ensure efficient production capacity, the unloading and processing rates are both regarded as maximum limits in the LSCOSP. Accordingly, the objective function can be simplified to minimize the changeover cost and is presented as follows:
\begin{align}
	\scalebox{1}{${F = {\rm{ }}\omega \sum\limits_{u\in U, n \in N} {CO_{u,n}},}$} \label{objFun}
\end{align}
where $\omega$ is the cost coefficient for the single changeover. Eq. \eqref{objFun} indicates that the changeover cost is incurred by switching feed among different tanks and switching operating modes in CDUs.

In this way, the MINLP formulation for the LSCOSP can be described as follows:
\begin{align}
	\left\{ \begin{array}{l}
		\min \,\,\,F\\
		s.t. \quad \rm{Eqs.} \eqref{a-1}-\eqref{b-2},\\
		\qquad \,\,\rm{Eqs.} (S.1)-(S.23).
	\end{array} \right.
\end{align}

\subsection{Illustrative Instance}

We provide a simple example to illustrate the crude oil scheduling process.
This example involves a single vessel ($V_1$), four crude tanks ($T_1,T_2,T_3,T_4$), three types of crude oil ($C_1,C_2,C_3$), and one CDU ($CDU_1$) capable of producing two types of residue ($R_1,R_2$) in a refinery. The scheduling horizon spans three days. Table \ref{example} lists crucial parameters, including the initial quantities of crude oil ($IV$) of the vessel and its arrival time ($AD$). The table also presents the initial inventory of tanks ($ITC$), the sulfur content ($P_{sul}$), and the residue yield ($Yld$) of each crude type. Additionally, the last row in this table outlines the permissible range of crude oil types for processing with each residue type. To be specific, vessel $V_1$ carries 60 kt of crude $C_2$ and is scheduled to arrive on the second day. The initial tank inventory comprises 50 kt of crude $C_3$, 30 kt of crude $C_1$, 20 kt of crude $C_2$, and 20 kt of crude $C_1$ for tanks $T_1$-$T_4$, respectively. The sulfur contents of crude $C_1$-$C_3$ are 0.5\%, 1.1\%, and 2.7\%, respectively, with corresponding residue yields of 33.64\%, 13.65\%, and 29.65\%. Due to the unique design regulations of each CDU, they can produce different types and quantities of residue by blending various crude oils. In the given example, CDU1 can produce residue $R_1$ by utilizing crude $C_1$-$C_3$, and it can also generate residue type $R_2$ by blending crude $C_1$ and $C_2$.

Fig. \ref{exFig} illustrates a feasible solution with two changeovers over three days for the above scenario, which meets all constraints described in Section \ref{formulation}. More specifically, Fig. \ref{exFig}(a) clearly demonstrates the feasibility of the operating constraints. Figs. \ref{exFig}(b) and \ref{exFig}(c) reflect the alteration of the feed status (e.g., flow rate and sulfur content) supplied to the CDU and the corresponding residue outputs during each changeover. It is evident that ensuring the feed status complies with specific boundary limitations and adjusting the residue processing mode for the CDU to meet the production demand are both critical for maintaining scheduling feasibility in terms of materials transfer.

However, it is important to note that the number of discrete variables and nonlinear terms significantly increases with the expansion of the scheduling horizon and the number of resources, such as tanks, crude types, CDUs, and product types. This expansion results in exponential time complexity and can even exceed the solving capacity of MINLP solvers. Given the advantages of EAs in solving large-scale and nonconvex problems, we propose a novel EA-based approach called DSEA/HR in the following section for solving the presented LSCOSP.

\begin{table}[!t]
	\setlength{\abovecaptionskip}{0.1em}
	\centering
	\caption{Numerical Example of the Studied Problem.}\label{example}
	\renewcommand{\arraystretch}{1.1}
	\resizebox{\linewidth}{!}{
		\begin{tabular}{p{2cm}<{\centering}p{6cm}<{\centering}}
			\toprule
			\multirow{2}{*}{\makecell[c]{Vessel \\($c|IV|AD$)}} & $V_1$\\
			\cline{2-2}
			~ & ($C_2 |$ 60 kt $|$ 2$^{nd}$ day)\\
			\hline \hline
	\end{tabular}}
	
	\resizebox{\linewidth}{!}{
		\begin{tabular}{p{2cm}<{\centering}p{1.3cm}<{\centering}p{1.3cm}<{\centering}p{1.3cm}<{\centering}p{1.3cm}<{\centering}}
			\multirow{2}{*}{\makecell[c]{Tank \\($c|ITC$)}} & $T_1$ & $T_2$ & $T_3$ & $T_4$ \\
			\cline{2-5}
			~ & ($C_3|$50\! kt) & ($C_1|$30\! kt) & ($C_2|$20\! kt) & ($C_1|$20\! kt) \\
			\hline \hline
	\end{tabular}}
	
	\resizebox{\linewidth}{!}{
		\begin{tabular}{p{2cm}<{\centering}p{1.75cm}<{\centering}p{1.75cm}<{\centering}p{1.75cm}<{\centering}}
			\multirow{2}{*}{\makecell[c]{Crude \\( $P_{sul}|Yld$)}} & $C_1$ & $C_2$ & $C_3$ \\
			\cline{2-4}
			~ & (0.5\%$|$33.64\%) & (1.1\%$|$13.65\%) & (2.7\%$|$29.65\%) \\
			\bottomrule
	\end{tabular}}
	
	\resizebox{\linewidth}{!}{
		\begin{tabular}{p{2cm}<{\centering}p{2.8cm}<{\centering}p{2.8cm}<{\centering}}
			\multirow{2}{*}{\makecell[c]{Residue \\ ${c}$}} & $RC_{R1}$ & $RC_{R2}$ \\
			\cline{2-3}
			~ & $\{C_1, C_2, C_3\}$ & $\{C_1, C_2\}$ \\
			\bottomrule
	\end{tabular}}
\end{table}

\begin{figure}[!t]
		\setlength{\abovecaptionskip}{-1em}
		\centering
		\includegraphics[scale= 0.55]{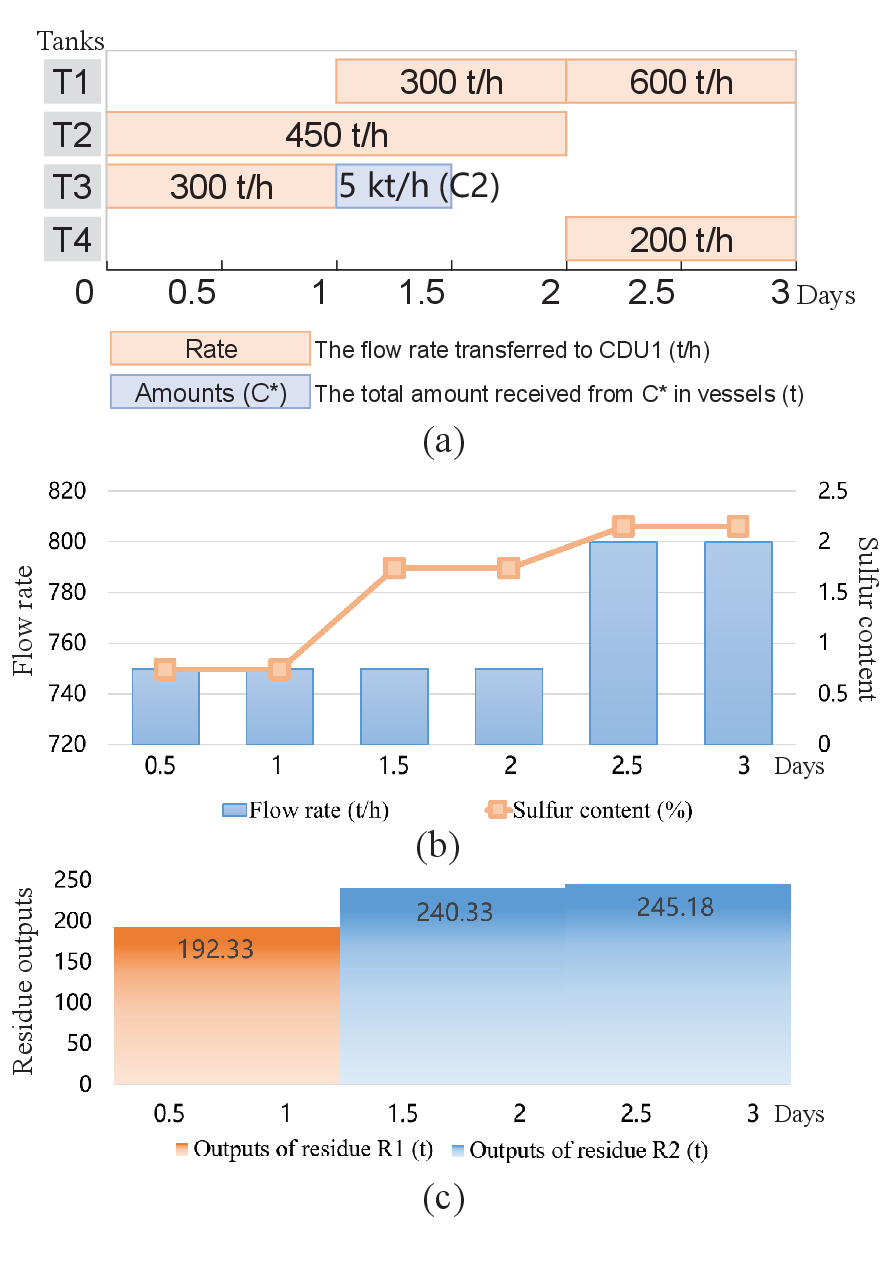}
		\caption{An illustration of an feasible schedule. (a) Gantt chart of a feasible solution. (b) Flow rate and sulfur content of feed to the CDU. (c) Outputs of two types of residue in the CDU.}
		\label{exFig}
		\vspace{-0.5em}
\end{figure}

\section{Dual-Stage Evolutionary Optimization for LSCOSPs} \label{section 3}
For large-scale problems, metaheuristic algorithms are popularly used since the above MINLP formulations are difficult to converge.
This section introduces a dual-stage evolutionary algorithm called DSEA/HR to solve the LSCOSP rather than directly solving the large-scale MINLP model. The core of the proposed DSEA/HR lies in its dual-stage search mechanism, which includes a global search and a local refinement stage. According to the diverse search characteristics of these two stages, we employ the competitive swarm optimizer (CSO) \cite{Cheng2015a} and the composite differential evolution (CoDE) \cite{Wang2011} optimizer as the fundamental optimizers in global and local optimization, respectively. It is worth noting that the key performance improvement stems from the dual-stage search framework itself, not limited to any specific optimizer.
The implementation details of the proposed DSEA/HR are elaborated below.

\subsection{Encoding and decoding schemes}
\begin{figure}[!t]
	\centering
	\includegraphics[scale= .3]{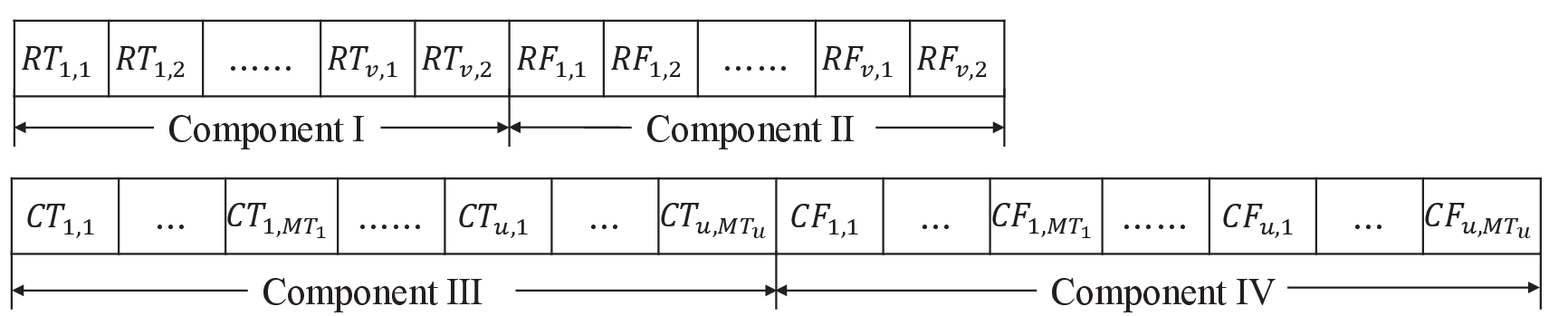}
	\caption{Representation of encoding for each period.}
	\label{encoding}
\end{figure}

In solving the LSCOSP, the encoding scheme is essential for EAs. A potential solution should reflect the operating target tanks and their flow rate. To this end, the structure of the encoding scheme for one period is comprised of four components: (1) receiving tanks, (2) receiving flow rate, (3) charging tanks, and (4) charging flow rate. They are illustrated in Fig. \ref{encoding}. In Component \uppercase\expandafter{\romannumeral1}, every two bits ${RT}_{v,1}$ and ${RT}_{v,2}$ denote receiving tanks. Similarly, in Component \uppercase\expandafter{\romannumeral2}, every two bits ${RF}_{v,1}$ and ${RF}_{v,2}$ indicate receiving flow rate. The size of them is $2|V|$ in one period, where $|V|$ is the total number of vessels. Moreover, in Components \uppercase\expandafter{\romannumeral3} and \uppercase\expandafter{\romannumeral4}, every single bit represents a charging tank and its charging flow, respectively. The size of them is the sum of ${MT}_u$ for all CDUs.

Based on the above encoding, a complete schedule result can be mapped clearly. In every period, the unloading tanks and their amounts for each vessel are respectively determined by Component \uppercase\expandafter{\romannumeral1} and \uppercase\expandafter{\romannumeral2} from left to right one by one. Note that the effectiveness of bits depends on whether the vessel $v$ is unloading. The charging tanks and their amounts for $u$-th CDU are respectively indicated by \{$CT_{u,1}$, ... ,$CT_{u,FT_u}$\} and \{$CF_{u,1}$, ... ,$CF_{u,{MT}_u}$\} represented in Component  \uppercase\expandafter{\romannumeral3} and \uppercase\expandafter{\romannumeral4}. To guarantee the sustainability of activities in the entire scheduling horizon, each vessel can unload from two tanks in a specific order during each period. Additionally, the duration of charging operations must be in integer periods, implying switches are only permitted at the beginning of the next period.

\subsection{Properties of LSCOSPs}
In the LSCOSPs, each type of crude oil is limited to producing a specific range of residue types. For instance, let $(C_{\Delta}, R_\Delta)$ denote the set of crude types in a tank ($\Delta\!=\!t$) or a vessel ($\Delta\!=\!v$) and their corresponding residue types that can be produced, respectively. If $(C_1, R_1)$ and $(C_2, R_2)$ satisfy the conditions $R_1\! \!=\!\! \{1,2\}$ and $R_2 \! \!=\! \! \{2,3\}$, then the blended crude $C_{blend} \! =\!  \{C_1 \cup C_2\}$ can only produce the residue type $R_{blend}  \!=\! \{1,2\}\cap\{2,3\} = \{2\}$. According to this principle, two properties of the LSCOSPs can be derived and used as valuable knowledge to enhance search efficiency.

\textit{Property 1:} When assigning discharging tanks, it is important to avoid significant disparities in the producible residue types between the crude oil in the discharging tank and that in the vessel.

\textit{Property 2:} After selecting charging tanks for CDUs, it is crucial to ensure the diversity of residue types produced by other storage tanks.

Notably, formulating the above properties in terms of equations is complicated. Therefore, we propose several heuristic rules that represent the knowledge as probabilities and incorporate them into the search process.

\subsection{Population initialization with heuristic rules} \label{hr}
In order to improve convergence for solving LSCOSPs, two heuristic rules based on empirical knowledge are utilized in the population initialization of DSEA/HR. Accordingly, Rules \uppercase\expandafter{\romannumeral1} and \uppercase\expandafter{\romannumeral2} are detailed below, followed by the procedure of population initialization shown in Algorithm \ref{alg-init}.

Rule \uppercase\expandafter{\romannumeral1} is devoted to the assignment of discharging tanks. According to Property 1, randomly blending different types of crude may easily lead to an infeasible schedule since different types of residue are only produced by specific crudes. In this work, tanks are given different priorities for discharging. Firstly, the highest priority is assigned to the empty tank. If there is no empty tank, we define the similarity between the unloaded crude and the mixture in the tank, formulated in Eq. \eqref{4-1}. Then the tanks with the highest similarity are chosen because the tanks with higher similarity own a wider processable range of residue types than others after mixing the unloaded crude. Finally, the tank with the largest available capacity is selected to receive among the tanks with the highest similarity.
\begin{equation}
	\scalebox{0.9}{${R_{v,c,t}} = \left| {\bigcap\limits_{c \in \left\{ {\mathop {UJ}\nolimits_v ,\mathop {BJ}\nolimits_t } \right\}} {CP}_c} \right|,\forall v,t,$} \label{4-1}
\end{equation}
where ${R_{v,c,t}}$ represents similarity between crude types in the tank \textit{t} and the crude \textit{c} to be unloaded from vessel $v$; ${CP}_c$ denotes the range of types in which the residue produced by crude \textit{c}; ${UJ}_v$ and ${BJ}_t$ are the types of crude unloaded from vessel \textit{v} and stored in tank \textit{t}, respectively.

\begin{algorithm}[!t]
	\footnotesize
	\caption{Population initialization.}
	\label{alg-init}
	\begin{algorithmic}[1]
		\REQUIRE  {$M$ (population size), problem parameters}
		\ENSURE  {$P(M)$ (initialized population)}
		\FOR{$m=1 \to M$}
		\STATE Randomly initialize $\{CF_{1,1}, \cdots, CF_{u,{MT}_u}\}$ as charging amount for each CDU;			
		\STATE Generate $\{CT_{1,1}, \cdots, CT_{u,{MT}_u}\}$ as the charging tanks for each CDU $\leftarrow$ Rule \uppercase\expandafter{\romannumeral2} and $\{CF_{1,1}, \cdots, CF_{u,{MT}_u}\}$;			
		\STATE Generate $\{RT_{1,1}, \cdots, RT_{v,2}\}$ as receiving tanks for each vessel $\leftarrow$ Rule \uppercase\expandafter{\romannumeral1} and initial capacity of tanks;			
		\STATE Calculate $\{RF_{1,1}, \cdots, RF_{v,2}\}$ as unloading amount based on  maximum capacity of $\{RT_{1,1}, \cdots, RT_{v,2}\}$;			
		\STATE Add initial solution $m$ into population $P(M)$;
		\ENDFOR
	\end{algorithmic}
\end{algorithm}

Rule \uppercase\expandafter{\romannumeral2} aims to generate a group of charging tanks for each CDU. Typically, the crudes in a tank can produce multiple types of residue. According to Property 2, it is promising to satisfy the inventory of residue if the tanks that can produce a variety of residue types are reserved. In this way, the probability of selection for the charging tank is presented in Eq. \eqref{4-2}. The tank producing fewer types of residue is preferred to be selected, ensuring a wide feasible region in the later periods.
\begin{equation}
		\scalebox{0.98}{$Prob_t = \frac{{\frac{1}{{\left| {\bigcap\limits_{c \in {BJ}_t} {CP}_c } \right|}}}}{{\sum\limits_{t' \in AT} {\frac{1}{{\left| {\bigcap\limits_{c \in {BJ}_{t'}} {CP}_c } \right|}}} }}, t \in AT,$}
	\label{4-2}
\end{equation}
where $AT$ indicates a set of crude tanks whose level is not fall below the lower bound after they charge CDU with amount specified by $CF$.

\subsection{Constraints handling and fitness evaluation}
In this paper, the LSCOSP is formulated as a large-scale constrained optimization problem:
\begin{equation}
	\begin{aligned}\label{4-8}
		\mathop {\min }\limits_x \quad &f(x) \\
		s.t. \quad &x=(x_1,\cdots,x_D) \in \Omega \\
		&{g_i}(x) \le 0, i = 1, \cdots ,p \\
		&{h_i}(x) = 0, i = p+1, \cdots ,q 		
	\end{aligned},
\end{equation}
where $x$ is a solution including $D$ decision variables; $\Omega$ is the decision space; ${g_i}(x)$ and ${h_i}(x)$ are the $i$-th inequality and $(i-p)$-th equality constraint, respectively; $f(x)$ is the objective value of solution $x$. Eqs. \eqref{4-9} and \eqref{4-10} respectively define the degree of violation and the binary flag for $i$-th constraint.
\begin{align}
	\begin{split}\label{4-9}
		CV_i(x) &=\left\{\!\!\! {\begin{array}{*{20}{l}}
				\max (0,{g_i}(x)), &\!\!\!\!if \, i \le p \\
				\max (0,\left| {{h_i}(x)} \right|), &\!\!\!\!otherwise
		\end{array}} \right.\!\!\!,i = 1, \ldots, q,
	\end{split}
\end{align}
\begin{align}
	\begin{split}\label{4-10}
		CVN_i(x) &= \left\{ {\begin{array}{*{20}{l}}
				1, &CV_i(x) > 0 \\
				0, &otherwise
		\end{array}} \right.,i = 1, \ldots, q,
	\end{split}	
\end{align}

Accordingly, the number and the degree of constraint violations are calculated by Eqs. \eqref{4-11} and \eqref{4-12}, respectively. Meanwhile, the objective value of individual $x$ is obtained from $f(x)$. In this way, the fitness of individual $x$ can be denoted as $(CVN(x),CV(x),f(x))$.
\begin{gather}
	\scalebox{0.9}{$CVN(x) = \sum\nolimits_{i = 1}^q {C{VN_i}(x)},$}
	 \label{4-11} \\
	\scalebox{0.9}{$CV(x) = \sum\nolimits_{i = 1}^q {C{V_i}(x)}.$} \label{4-12} 	
\end{gather}

Further, the feasibility-based criterion \cite{Deb2002} is used in the solution selection to accelerate the convergence of the population to feasible regions. To be specific, the fitness of solution $x$ is better than another solution $y$ if the following conditions hold:
\begin{align}
	\begin{split}
		\left\{\!\! {\begin{array}{*{20}{l}}
				\scalebox{0.9}{$\!f(x) \!<\! f(y),$} & \scalebox{0.9}{$\!\!if \, CVN(x),CVN(y) \!= 0$} \\
				\scalebox{0.9}{$\!CV(x) \!<\! CV(y),$} & \scalebox{0.9}{$\!\!if \, CVN(x) \!= CVN(y)$} \\
				\scalebox{0.9}{$\!f(x) \!<\! f(y),$} & \scalebox{0.9}{$\!\!if \, CV(x) \!= CV(y)$} \\	
				\scalebox{0.9}{$\!CVN(x) \!<\! CVN(y),$} & \scalebox{0.9}{$\!\!otherwise$} 		
		\end{array}} \right..
	\end{split}
\end{align}
The above selection operation is adopted into both the global and local search stages.

\subsection{Global search stage for LSCOSPs} \label{glo}
In the dual-stage framework of the proposed DSEA/HR, the global search serves as the first stage. Its goal is to evolve populations within a mixed decision space that involves large-scale discrete and continuous variables. Given that the performance of conventional optimizers, such as the particle swarm optimizer (PSO) \cite{Kennedy1995}, seriously degrades as the dimension of the search space increases. It is more appropriate to embed optimizers specifically designed for large-scale global optimization problems into the global search stage of DSEA/HR. To this end, we have adopted CSO \cite{Cheng2015a} as the basic optimizer for the global search stage in this paper. CSO is well-suited for solving widely used high-dimension problems due to its novel competition mechanism.

In CSO, neither the particle's best position nor the global best position from the basic PSO is used. Instead, the particles are randomly grouped in pairs. Then the particles from the same pair compete with each other. The particle with better fitness will be selected to proceed to the next generation. The other one will be updated by the modified velocity and position learned from the winner in this competition. The updated strategies are presented in Eqs. \eqref{4-5} and \eqref{4-6}.
\begin{gather}
	\begin{split}
		\scalebox{0.9}{${V_{l,n}}(g + 1) =$} &\scalebox{0.9}{${R_1}(n,g){V_{l,n}}(g)$} \\ & \scalebox{0.9}{$+ {R_2}(n,g)({X_{w,n}}(g) - {X_{l,n}}(g))$} \\&  \scalebox{0.9}{$+ \varphi {R_3}(n,g)(\overline {{X_n}}(g) - {X_{l,n}}(g)),$} \label{4-5}
	\end{split} \\
	\scalebox{0.9}{${X_{l,n}}(g + 1) = {X_{l,n}}(g) + {V_{l,n}}(g + 1),$} \label{4-6}
\end{gather}
where $V_{l,n}(g + 1)$ and $X_{l,n}(g + 1)$ denote the velocity and position of the loser in the $n$-th round of the competition in generation $t$+1, respectively; ${R_1}(n,g)$, ${R_2}(n,g)$, ${R_3}(n,g)$ are vectors randomly generated between 0 to 1 after the \textit{n}-th competition in generation \textit{g}; $\overline {{X_n}}(g)$ is the mean value vector of all particles in generation \textit{g} for each dimension; $\varphi$ is a control parameter, which is set to zero as recommended in \cite{Cheng2015a}.

The above strategies for updating velocity and position are used to generate the swarm of the next generation in the global search stage. Algorithm \ref{alg-cso} presents the procedure of the global search with CSO.

\begin{algorithm}[!t]
	\footnotesize
	\caption{Procedure of the global search.}
	\label{alg-cso}
	\begin{algorithmic}[1]
		\REQUIRE $P_{0}(M)$ (initialized population), maximum evaluation number
		\ENSURE $S_{CSO}(Z)$ (best solutions by CSO)
		\STATE $g=0$;
		\STATE /*Population evolution with CSO */
		\WHILE{the maximum evaluation number is not achieved}
		\STATE Calculate the $fitness$ of each particle $X_i(g)$ in $P_g(M)$ using \eqref{4-8}-\eqref{4-12};
		\STATE $A = P_{g}(M)$, $P_{g+1} = \varnothing$;
		\WHILE{$A \ne \varnothing $}
		\STATE Randomly select two particles $X_1(g)$, $X_2(g)$ from $A$;
		\IF{$fitness(X_1(g))>fitness(X_2(g))$}
		\STATE $X_w(g)=X_1(g)$, $X_l(g)=X_2(g)$;
		\ELSE
		\STATE $X_w(g)=X_2(g)$, $X_w(g)=X_1(g)$;
		\ENDIF
		\STATE $P_{g+1}(M) \leftarrow$  $X_w(g)$;
		\STATE Update $X_l(g)$ using \eqref{4-5} and \eqref{4-6};
		\STATE $P_{g+1}(M) \leftarrow$ Updated $X_l(g)$;
		\STATE Remove $X_1(g)$, $X_2(g)$ from $A$;
		\ENDWHILE	
		\STATE $g = g+1$;
		\ENDWHILE
		\STATE $S_{CSO}(Z) \leftarrow $ Select solutions with the best fitness from $P_g(M)$;
	\end{algorithmic}
\end{algorithm}

\begin{remark}
	It is worth mentioning that the proposed dual-stage evolution framework is flexible and can accommodate any optimizer. Thus, the choice of the employed CSO for the proposed DSEA/HR is not unique. While the optimizer for the global stage does not play a decisive role in terms of performance improvement, using an optimizer tailored for large-scale global optimization is helpful in improving the algorithm's efficacy. This advantage is also demonstrated by the detailed experimental comparison provided in Section \ref{Comparison2}.
\end{remark}

\subsection{Repair strategy based on local refinement} \label{loc}

After optimizing the global search stage for the LSCOSPs, the discrete variables representing discharging and charging tanks in the elite solutions are expected to converge, aided by the guidance of heuristic rules. However, finding a feasible solution in the global search stage remains challenging due to the complexity of the mixed decision space. To mitigate this, we have developed a local refinement stage with a problem-specific repair strategy for handling infeasible solutions.

To enhance the exploitation capability of DSEA/HR, we have focused on optimizing continuous variables. We have introduced a repair strategy that fixes the discrete variables, effectively transforming the mixed-variable problem into a continuous-variable one. This strategy, focusing on local variables (continuous variables), is termed \textit{local refinement}.
Specifically, during the local refinement stage, the values of discrete variables inherit the results obtained from the global search, while the charging flow represented by the continuous variable will be optimized in the EA-based framework. Note that the EA-based optimizer cannot be replaced by any deterministic nonlinear solver. This is because the preservation of the diversity of feasible solutions is essential in solving the scheduling problem effectively.

\begin{algorithm}[!t]
	\caption{Procedure of the local refinement.}
	\label{alg-de}
	\footnotesize
	\begin{algorithmic}[1]
		\REQUIRE $S_{CSO}(Z)$ (best solutions by CSO), $H$ (population size), maximum evaluation number
		\ENSURE  $S_{best}$ (final solution)
		\IF{$S_{CSO}(Z)$ violate processing constraints}
		\STATE /* Population initialization for local refinement */
		\STATE $i = 1$, $ n = 0 $, $g=0$;
		\FOR{$i=1 \to H$}
		\STATE Randomly initialize $\{CF_{1,1}, \cdots, CF_{u,{MT}_u}\}$ as charging amount for each CDU for individual $i$ ;
		\STATE Duplicate charging tanks for each CDU $\{CT_{1,1}, \cdots, CT_{u,{MT}_u}\}$ from the particle $n$ to the individual $i$;
		\STATE Add individual $i$ into $Q_g$;
		\STATE $n=n+1$;
		\IF {$n>Z$}
		\STATE $n=1$;
		\ENDIF
		\ENDFOR
		\STATE Calculate the $fitness$ of each individual $X(g)$ in $Q_g(H)$ using \eqref{4-8}-\eqref{4-12};
		\STATE /*Population evolution with CoDE */
		\WHILE{the maximum evaluation number is not achieved}
		\STATE $Q_{g+1} = \varnothing$;
		\FOR{$i=1\to n$}
		\STATE Generate trail vectors of $X(g)$ by the generation strategies and random control parameters in CoDE;
		\STATE Evaluate trail vectors and select them with $X(g)$ using \eqref{4-8}-\eqref{4-12};
		\STATE $Q_{g+1}$ $\leftarrow$ Update charging amount for each CDU $\{CF_{1,1}, \cdots, CF_{u,{MT}_u}\}$ of $X(g)$;
		\ENDFOR
		\STATE $g = g+1$;
		\ENDWHILE
		\STATE $ S_{best} \leftarrow $ Select solutions with the best fitness from $Q_g(H)$;
		\ELSE
		\STATE $ S_{best} \leftarrow S_{CSO}(Z)$;
		\ENDIF
	\end{algorithmic}
\end{algorithm}

With respect to the choice of the optimizer, it is hopeful for local search operators to enhance the exploitation ability over continuous spaces. With this information, we employ CoDE \cite{Wang2011} as an exemplary optimizer to enhance the exploitation ability in the local refinement stage.
In CoDE, the control parameters and trial vector generation strategies are randomly selected from a candidate pool. The candidate pool of strategies and parameters can be adjusted to suit different problem-solving scenarios. Algorithm \ref{alg-de} lists the procedure of the local refinement using CoDE for repairing infeasible solutions.
In Lines 3-12, the initial population is composed of individuals represented by continuous variables, while the discrete variables remain the same as elite solutions from the global stage. The population for the local refinement is evolved using CoDE operators to find feasible solutions, as shown in Lines 15-23 of Algorithm \ref{alg-de}.

\textit{Remark 2:} It should be noted that the optimization scopes of the global search stage and the local refinement stage are distinct. The global search stage emphasizes the optimization of discrete variables, assisted by heuristic rules, while the local refinement stage focuses on optimizing the continuous variables of potential solutions. The integration of these two stages is expected to enhance the performance of the proposed DSEA/HR in mixed decision spaces. Furthermore, the selection of the optimizer for the local refinement stage is not confined to a single option. An optimizer with a robust local exploitation ability, serving the purpose of effective local refinement, would facilitate the improvement of solutions towards more promising regions.

\begin{figure}[!t]
	\centering
	\includegraphics[scale= .55]{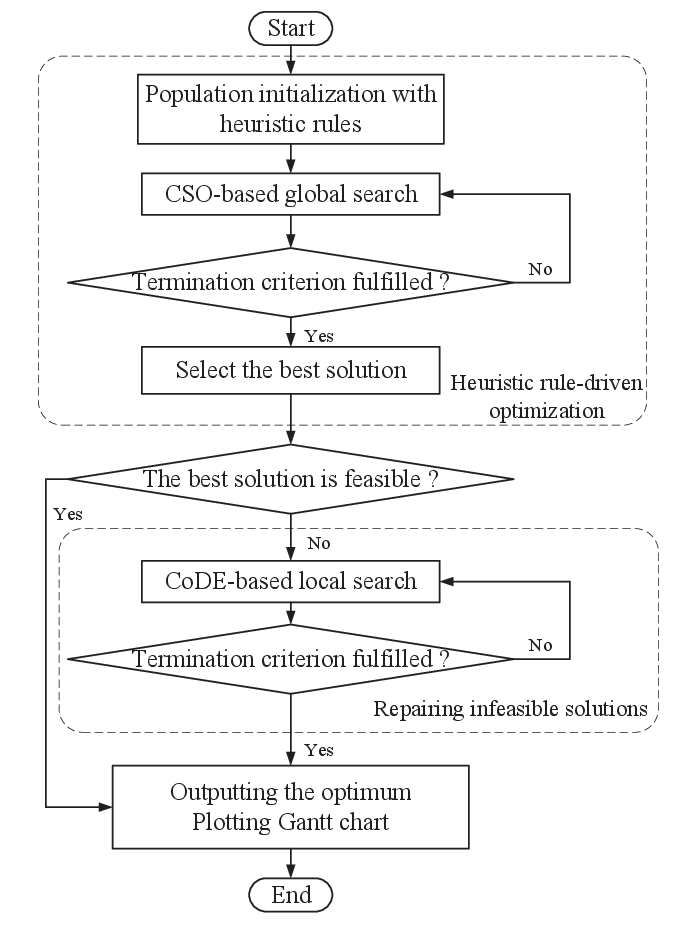}
	\caption{The flowchart of DSEA/HR for solving LSCOSPs.}
	\label{flowChart}
\end{figure}

\subsection{Framework of DSEA/HR}
The flowchart of DSEA/HR is illustrated in Fig. \ref{flowChart}, which consists of two fundamental building modules: global search and local refinement. The initial population is generated based on the heuristic rules. In the process of evolution, two effective operators (i.e., CSO and CoDE) are adopted in global search and local refinement, respectively. Individual fitness is evaluated by the designed criterion before the selection. If the global search stage fails to obtain a feasible solution, the population can be repaired by the local refinement stage.

\section{Case Study} \label{section 4}

\subsection{Experimental Settings} \label{ex}
 The effectiveness of the proposed DSEA/HR on LSCOSPs is demonstrated by the large-scale cases from a real-world marine-access refinery. The refinery has two berths, twenty-one tanks (T1-T21), three CDUs (CDU1-CDU3), and four types of residue tanks (R1-R4). The scheduling horizon is ten days (20 periods) since the short-term scheduling of the refinery is carried out three times a month. For the same refinery, the complexity of the scenarios is influenced by certain conditions, such as the crude oil parcels to receive, initial tank inventory, and the properties of crude oil. Three cases are tested in the case study. Case 1 considers several new types of parcels that are different from the initial component in each tank. The challenge of this case is that unloading operations may significantly affect the stability of crude components in tanks. Case 2 and Case 3 have a large number of low-quality crudes. In particular, more than half of the crudes are high sulfur content or heavy density in Case 3. Proper blending is crucial to satisfy the constraints of properties and yields. The details of all the cases are listed in Section S.II of the supplementary file.

 \label{group}
Three groups of comparison experiments were designed as follows.

\emph{Group 1:} Comparison between DSEA/HR and state-of-the-art commercial MINLP optimization solvers, including ANTIGONE \cite{Misener2014}, SCIP \cite{Achterberg2009}, and SHOT \cite{Lundell2022}.

\emph{Group 2:} Comparison between DSEA/HR and state-of-the-art metaheuristics. This includes DSEA/HR-PSO, where CSO is replaced by PSO \cite{Kennedy1995} in the global search stage of DSEA/HR, as well as CSO \cite{Cheng2015a}, the efficient recursive differential grouping (ERDG) \cite{Yang2021}, the self-adaptive differential evolution with multi-trajectory search (SaDE-MMTS) \cite{Zhao2011}, and two metaheuristics for crude oil scheduling mentioned in Section \ref{section 1}, namely SAGA \cite{Ramteke2012} and COSO-GA for single objective \cite{Hou2017}.

\emph{Group 3:} Comparison between DSEA/HR and its two variants, including DSEA/HR without the incorporation of heuristic rules (denoted by DSEA/HR-V1) and DSEA/HR without local refinement (denoted by DSEA/HR-V2).

All metaheuristics were implemented with MATLAB 2020b, while the MINLP model for LSCOSP was programmed and solved in GAMS version 33.1. In comparison with deterministic approaches, the performance of algorithms was evaluated using objective values and CPU times as metrics. The mean, standard deviation (std), and feasible rate (FR) \cite{Liu2019} were used as indicators to evaluate the optimality and stability of experimental results compared with metaheuristics.

To ensure fair comparisons, each case was optimized using DSEA/HR or other metaheuristics over 20 independent runs. The Wilcoxon's rank-sum test at a 0.05 significance level was employed to statistically compare the algorithm performance. In addition, all experiments were executed on Windows 10 with Intel(R) Xeon(R) Gold 6256 CPU @ 3.60 GHz and 16.0 GB RAM.

\subsection{Parameter Settings} \label{Parameter}
\subsubsection{Operators}
The proposed DSEA/HR integrates the CSO in the global search stage and the CoDE in the local refinement stage. To demonstrate the simplicity and adaptability of our approach, the control parameters in CSO  (i.e., $\varphi$) and CoDE (i.e., scaling factor and crossover rate) are set based on their original papers' recommendations \cite{Cheng2015a,Wang2011}.

\subsubsection{Swarm and population size} \label{R1.4.1}
The swarm size of CSO is set to 100 according to the recommendation in \cite{Cheng2015a}, while the population size of CoDE requires determination, as setting details are not provided in \cite{Wang2011}.
Table \ref{sensitive} presents a sensitivity analysis on different population sizes ($PS$). The performance of DSEA/HR using $PS\!\!=\!\!60$ outperforms than those using $PS\!\!=\!\!30$ and $PS\!\!=\!\!90$ in three cases. Consequently, we set $PS$ as $60$ in this paper.

\begin{table}[!t]
	\centering
	\large
	\caption{Indicator Results Obtained by DSEA/HR that Adopts Different Setting Values for $PS$ after 20 Runs.} \label{sensitive}
	\renewcommand{\arraystretch}{1.2}
	\resizebox{\linewidth}{!}{
		\begin{threeparttable}
			\begin{tabular}{|c|c|c|c|c|c|c|}
				\hline
				\multirow{2}{*}{Case} & \multicolumn{2}{c|}{PS=30} &  \multicolumn{2}{c|}{PS=60} & \multicolumn{2}{c|}{PS=90} \\
				\cline{2-7}
				~ & mean$\pm$std & FR & mean$\pm$std & FR & mean$\pm$std & FR \\
				\hline
				Case 1 & $13.7500\pm2.2913$ & $70\%$ & \boldmath{$12.2500\pm1.6819$}\tnote{*} & \boldmath{$95\%$} & $14.5000\pm2.7434$ & $80\%$ \\
				\hline
				Case 2 & $14.9000\pm4.3758$ & $60\%$ & \boldmath{$11.3000\pm1.5927$}\tnote{*} & \boldmath{$90\%$} & $13.2000\pm3.3182$ & $50\%$ \\
				\hline
				Case 3 & $25.3000\pm2.2501$ & $75\%$ & \boldmath{$23.4500\pm2.4382$}\tnote{*} & \boldmath{$80\%$} & $26.1000\pm2.7701$ & $60\%$ \\
				\hline
			\end{tabular}
			The best results are shown in boldface. `*' indicates that the result is significant better than the peer algorithm at a 0.05 level by the Wilcoxon's rank-sum test.
		\end{threeparttable}
	}
\end{table}

\subsubsection{Number of function evaluations}
The total number of fitness evaluations of the populations is used as the termination criterion and set to a sufficiently large value to ensure convergence. Specifically, the number of fitness evaluations for the two search stages of DSEA/HR is set as $1 \times 10^5$ and $3 \times 10^4$, respectively.

\subsection{Comparisons with other state-of-the-art algorithms} \label{5.2}
In this subsection, we first verify the effectiveness of DSEA/HR with traditional deterministic methods. Then we compare the performance of DSEA/HR with metaheuristic algorithms tailored for large-scale optimization problems.

\subsubsection{\textbf{Comparisons with deterministic algorithms}}
The scale of variables in the LSCOSP is related to the following product of several factors, i.e., the number of unloading crude types $\times$ tanks $\times$ periods, the number of tanks $\times$ CDUs $\times$ periods, and the number of CDUs $\times$ residue types $\times$ periods. Apparently, the scale of the problem is considerable with the increase in the scheduling horizon. In order to investigate the effectiveness of the proposed approach for large-scale optimization, we extend each of the above cases to two examples with different lengths of the horizon, in which one implements three-day scheduling, while the other carries out the scheduling with ten days. Table \ref{size} lists the problem scale of each example.
\begin{table*}[!t]
	\centering
	\caption{Problems scale for MINLP formulation}\label{size}
	\renewcommand{\arraystretch}{1.15}
	\resizebox{\linewidth}{!}{
		\begin{tabular}{|c|c|c|c|c|c|c|c|}
			\hline
			Case & Example & \makecell[l]{Parcels $\times$ Tanks $\times$ Crude types \\ $\times$ CDUs $\times$ Residue types}  & Horizon (periods) & Total var. & Binary var. & Total cons. & Nonlinear cons. \\
			\hline
			\multirow{2.5}{*}{Case 1} & Ex.1A & {$1 \times 19 \times 14 \times 3 \times 4$}  & 3 days (6) & 9948 & 538 & 19687 & 11172 \\
			\cline{2-8}
			~ & Ex.1B & {$6 \times 19 \times 14 \times 3 \times 4$} & 10 days (20) & 36960 & 3660 & 79942 & 40964\\
			\hline
			\multirow{2.5}{*}{Case 2} & Ex.2A & {$ 1 \times 21 \times 13 \times 3 \times$ 4}  & 3 days (6) & 10200 & 534 & 20062 & 11466 \\
			\cline{2-8}
			~ & Ex.2B & {$ 4 \times 21 \times 13 \times 3 \times$ 4}  & 10 days (20) & 35920 & 2740 & 75843 & 42042 \\
			\hline
			\multirow{2.5}{*}{Case 3} & Ex.3A & {$2 \times 21 \times 11 \times 3 \times 4$}  & 3 days (6) & 8928 & 546 & 17836 & 9702 \\
			\cline{2-8}
			~ & Ex.3B & {$8 \times 21 \times 11 \times 3 \times 4$}  & 10 days (20) & 33760 & 3820 & 74412 & 35574\\
			\hline
	\end{tabular}	}
\end{table*}

\begin{table*}[!t]
	\centering
	\caption{Comparison of MINLP Algorithms and DSEA/HR Results}\label{result-1}
	\renewcommand{\arraystretch}{1.15}
	\resizebox{\linewidth}{!}{
		\begin{threeparttable}
			\begin{tabular}{|c|c|c|c|c|c|c|c|c|}
				\hline
				\multirow{2.5}{*}{Example} & \multicolumn{2}{c|}{DSEA/HR} & \multicolumn{2}{c|}{ANTIGONE} & \multicolumn{2}{c|}{SCIP} & \multicolumn{2}{c|}{SHOT}  \\
				\cline{2-3}
				\cline{4-5}
				\cline{6-7}
				\cline{8-9}
				~ & Objective value & CPU time (s) & Objective value & CPU time (s) & Objective value & CPU time (s) & Objective value & CPU time (s) \\
				\hline
				Ex.1A & \textbf{0} & \textbf{15} & 0 & 110 & 0 & 234 & $-$ & $-$  \\
				\hline
				Ex.1B & \textbf{12} & \textbf{70} & $-$ & $-$ & $-$ & $-$ & $-$ & $-$  \\
				\hline
				Ex.2A & 1 & \textbf{14} & 0 & 540 & \textbf{0} & 474 & 10 & 43  \\
				\hline
				Ex.2B & \textbf{11} & \textbf{98} & $-$ & $-$ & $-$ & $-$ & $-$ & $-$  \\
				\hline
				Ex.3A & 2 & \textbf{19} & \textbf{0} & 2454 & $-$ & $-$ & $-$ & $-$  \\
				\hline
				Ex.3B & \textbf{23} & \textbf{286} & $-$ & $-$ & $-$ & $-$ & $-$ & $-$  \\
				\hline
			\end{tabular}
			The best results are shown in boldface. `$-$' indicates that no feasible solution was obtained within the allowable CPU time (3600 s).
		\end{threeparttable}
	}
\end{table*}

The proposed DSEA/HR is compared with three commercial MINLP solvers (i.e., ANTIGONE \cite{Misener2014}, SCIP \cite{Achterberg2009}, and SHOT \cite{Lundell2022}) on the above examples. These solvers integrate various state-of-the-art optimization technologies, such as nonlinear branch-and-bound \cite{Hooker2007}, extended supporting hyperplane \cite{Kronqvist2016}, and decomposition-based outer approximation \cite{Muts2020}.

Table \ref{result-1} presents the numerical results, in which we set the cost coefficient $\omega = 1 $ in the objective function for the convenience of calculation. For scheduling problems within three days (i.e., Examples 1A, 2A, and 3A), DSEA/HR and MINLP-based algorithms are able to find feasible solutions. Example 1A considers a network consisting of 19 tanks, three CDUs, and four types of residue tanks. In this example, initial inventory levels in crude tanks are high, facilitating the charging operation. It can be observed that DSEA/HR and two MINLP-based algorithms find the schedule without any changeover operations, while DSEA/HR has better performance on computing time. This is because MP-based methods require much time to deal with multiple equations, even in a case with a simple initial situation. In contrast, with the guidance of heuristic rules, the proposed EA-based approach can effortlessly search the feasible region on the condition of sufficient initial crude inventory. For Examples 2A and 3A, the number of tanks increases to 21, and more low-quality crude is contained in tanks. In Example 2A, the properties of crudes in eight tanks are high-sulfur or high-density, which exceeds the limits of the feed properties allowed by CDUs. Similarly, the properties of crude in five tanks are both high-sulfur and high-density in Example 3A. Apparently, the strict quality properties constraints can lead to a small feasible region, which affects the computing time and the accuracy of solutions. However, our approach obtains the approximate optimal results in a very short time, as shown in Examples 2A and 3A in Table \ref{result-1}. It is worth mentioning that the CPU time for DSEA/HR does not remarkably increase with the lower crude oil properties.

For Examples 1B-3B, the length of the scheduling horizon goes up to ten days. Except for more crude oil parcels needed to receive, the other conditions are the same as Examples 1A-3A. As observed in Table \ref{result-1}, the MINLP methods cannot find any feasible solution for Examples 1B-3B within the CPU time limit. In contrast, the proposed DSEA/HR identifies the feasible solution within an acceptable computing time. Overall, though our approach is inferior to the MP-based method in the optimality for general-scale problems, it is prospective to provide a feasible solution for larger problems via the mechanisms of heuristic rules and the repair strategy. In order to confirm this assertion, we will discuss the performance of the two mechanisms in Section \ref{eff}.

\subsubsection{\textbf{Comparisons with metaheuristic algorithms}} \label{Comparison2}
The proposed DSEA/HR is compared with six state-of-the-art metaheuristics (i.e., DSEA/HR-PSO, CSO, ERDG, SaDE-MMTS, SAGA, and COSO-GA) on LSCOSPs. Among these compared algorithms, DSEA/HR-PSO is a variant of DSEA/HR that utilizes PSO for the global search of the original DSEA/HR.
CSO is a simple and effective large-scale optimization algorithm that has also been adopted as the optimizer of the proposed DSEA/HR in the global search stage. ERDG, designed in the cooperative co-evolution framework, has been proposed to solve large-scale optimization problems with an efficient variable grouping strategy. SaDE-MMTS is a modification of the original SaDE, satisfying both global and local search requirements and demonstrating remarkable performance on the high-dimensional test functions up to 1000 dimensions.
SAGA and COSO-GA have both been developed to address crude oil scheduling problems: the former provides a graph-based structure for encoding, while the latter applies GA only to optimize integer variables.

\begin{table*}[!t]
	\centering
	\caption{Indicator Results Obtained by DSEA/HR and State-of-the-Art EAs after 20 Runs.}\label{result-2}
	\renewcommand{\arraystretch}{1.15}
	\resizebox{\linewidth}{!}{
		\begin{threeparttable}
			\begin{tabular}{|p{2.5cm}<{\centering}|c|c|c|c|c|c|}
				\hline
				\multirow{2.25}{*}{Algorithm}  & \multicolumn{2}{c|}{Case 1} & \multicolumn{2}{c|}{Case 2} & \multicolumn{2}{c|}{Case 3} \\
				\cline{2-7}
				~ & mean$\pm$std & FR & mean$\pm$std & FR & mean$\pm$std & FR \\
				\hline
				DSEA/HR  & \boldmath{$12.2500\pm1.6819$}\tnote{*}  & \boldmath{$ 95\%$} & \boldmath{$11.3000\pm1.5927$}\tnote{*}  &  \boldmath{$90\%$} & \boldmath{$23.4500\pm2.4382$}\tnote{*}  &  \boldmath{$80\%$} \\
				\hline
				DSEA/HR-PSO & $ 12.8000\pm 1.5079 $ & $ 95\%$ & $ 13.7500\pm 1.6162 $ & $ 75\%$ & $ 24.5500\pm 2.5021  $ & $ 70\%$ \\
				\hline
                CSO  & $ 14.3500\pm 2.7582 $ & $ 20\%$ & $ 15.5000\pm 2.6852 $ & $ 15\%$ & $ 26.1000\pm 2.2919  $ & $ 5\%$ \\
				\hline
				ERDG & $ 13.7000\pm 2.5152 $ & $ 10\%$ & $ 15.8000\pm 2.2384 $ & $ 10\%$ & $ 28.1000\pm 2.1250 $ & $ 0\%$ \\
				\hline
				SaDE-MMTS  & $ 15.7000\pm 1.6575 $ & $ 25\%$ & $ 14.4500\pm 2.1879 $ & $ 20\%$ & $ 26.6500\pm 1.4318 $ & $ 10\%$ \\
				\hline
				SAGA & $ 14.1500\pm2.5397  $ & $ 45\%$ & $ 14.1000\pm2.1002  $ & $ 15\%$ & $ 26.4500\pm 1.9595  $ & $ 10\%$  \\
				\hline
				COSO-GA & $ 15.2500\pm2.5726  $ & $ 50\%$ & $ 13.9000\pm1.6827  $ & $ 20\%$ & $ 25.5000 \pm 1.9057  $ & $ 20\%$  \\
				\hline
			\end{tabular}
				\scriptsize
				The best results are shown in boldface. `*' indicates that the result is significant better than the peer algorithm at a 0.05 level by the Wilcoxon's rank-sum test.
		\end{threeparttable}
	}
\end{table*}

From the results summarized in Table \ref{result-2}, it can be observed that DSEA/HR significantly outperforms the other comparison algorithms in term of both optimality and stability on all cases. Notably, three competitors (i.e., CSO, ERDG, and SaDE-MMTS) struggled to find feasible solutions, and ERDG even failed to find any feasible solution within 20 runs for the cases with the relatively small feasible regions. These outcomes emphasize that relying on a single optimizer is insufficient for real-world scheduling problems. The results obtained from SAGA and COSO-GA show that these two approaches, specifically designed for crude oil scheduling problems, do not exhibit significant performance when applied to solving large-scale instances. Additionally, the variant DSEA/HR-PSO is slightly worse than DSEA/HR in terms of optimality, while DSEA/HR-PSO is comparable to DSEA/HR in terms of stability. This comparison indicates that while the superior performance of DSEA/HR is mainly due to the dual-stage framework, the incorporation of powerful optimizers still contributes to identifying superior solutions.

The above experimental results demonstrate the significance of the dual-stage search framework in the proposed DSEA/HR, thereby verifying the main motivation of this article. Additionally, there exists an opportunity to further enhance the efficacy of the algorithm by incorporating several strategies. In the subsequent section, we will discuss the roles of kernel strategies designed for each stage of DSEA/HR, namely heuristic rules and the repair strategy.

\subsection{Effectiveness of the Components in DSEA/HR} \label{eff}

To investigate the effectiveness of two key components of DSEA/HR, namely the heuristic rules and the repair strategy, we adapt the original DSEA/HR into two variants. These variants are named DSEA/HR-V1 and DSEA/HR-V2. In DSEA/HR-V1, the heuristic rules for population initialization in the DSEA/HR algorithm are removed. As for DSEA/HR-V2, the repair strategy in the DSEA/HR algorithm is discarded.

\begin{table*}[!t]
		\centering
		\caption{Indicator Results Obtained by DSEA/HR and its Two Variants Algorithms after 20 Runs.}\label{result-3}
		\renewcommand{\arraystretch}{1.15}
		\resizebox{\linewidth}{!}{
			\begin{threeparttable}
				\begin{tabular}{|p{2.5cm}<{\centering}|c|c|c|c|c|c|}
					\hline
					\multirow{2.25}{*}{Algorithm}  & \multicolumn{2}{c|}{Case 1} & \multicolumn{2}{c|}{Case 2} & \multicolumn{2}{c|}{Case 3} \\
					\cline{2-7}
					~ & mean$\pm$std & FR & mean$\pm$std & FR & mean$\pm$std & FR \\
					\hline
					DSEA/HR  & \boldmath{$12.2500\pm1.6819$}\tnote{*}  & \boldmath{$ 95\%$} & \boldmath{$11.3000\pm1.5927$}\tnote{*}  &  \boldmath{$90\%$} & \boldmath{$23.4500\pm2.4382$}\tnote{*}  &  \boldmath{$80\%$} \\
					\hline
					DSEA/HR-V1  & $13.6500 \pm 1.8994 $ & $ 50\%$ & $14.9500 \pm 2.0641 $ & $ 40\%$ & $ 26.2000\pm1.8525  $ & $ 30\%$ \\
					\hline
					DSEA/HR-V2 & $ 13.1000\pm 1.6512 $ & $ 65\%$ & $ 12.5500\pm 1.5381  $ & $ 20\%$ & $ 25.2500\pm 1.6504 $ & $ 15\%$ \\
					\hline
				\end{tabular}
					\scriptsize
					The best results are shown in boldface. `*' indicates that the result is significant better than the peer algorithm at a 0.05 level by the Wilcoxon's rank-sum test.
			\end{threeparttable}
	}
\end{table*}

\begin{figure}[!t]
	\centering 	
	\includegraphics[scale= .21]{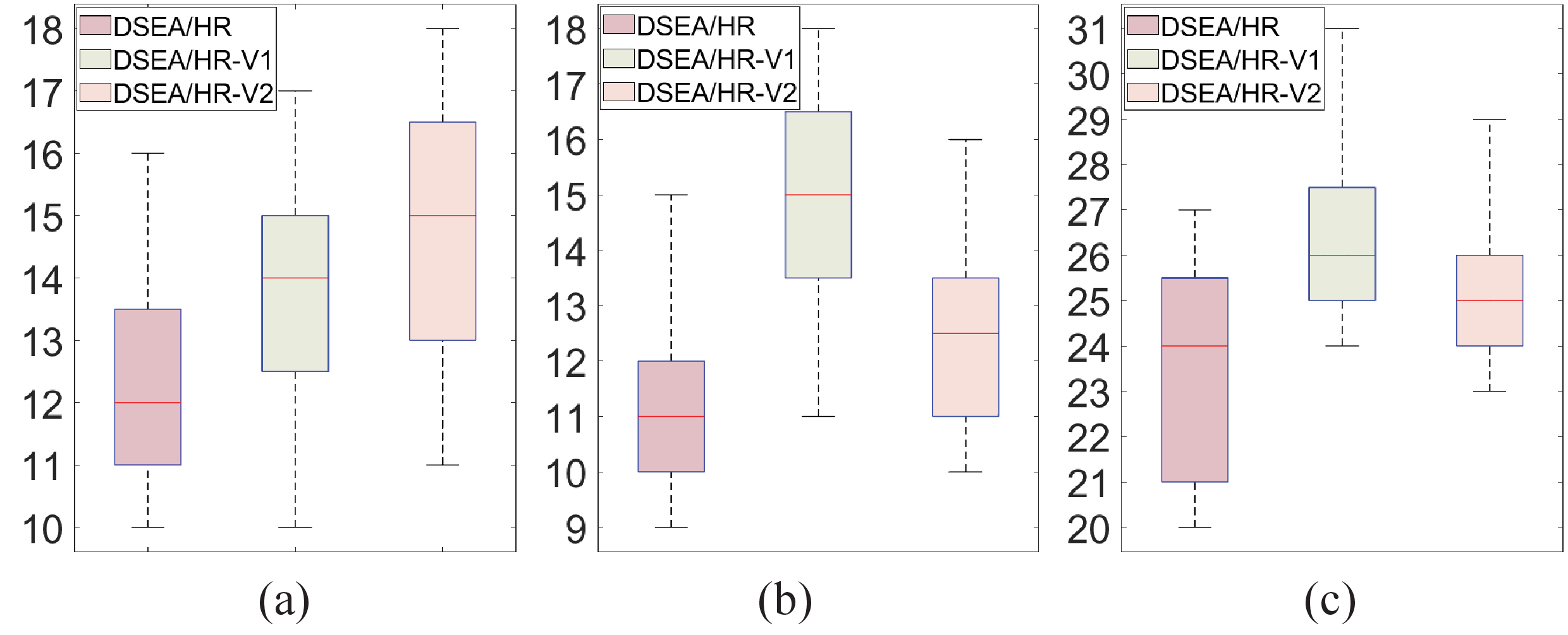}
	\caption{Box plots for the objective values by DSEA/HR, DSEA/HR-V1, and DSEA/HR-V2 in Cases 1-3. (a) Case 1. (b) Case 2. (c) Case 3.}\label{box-2}
\end{figure}

Table \ref{result-3} shows the statistical results obtained by DSEA/HR and the two variants. It can be observed that DSEA/HR significantly outperforms its two variants in all cases. Moreover, in the comparison of the two variants, due to the high inventory levels and the proper crude quality in Case 1, the guidance of heuristic rules is dominant. Thus, the performance of DSEA/HR-V2 with heuristic rules is better than DSEA/HR-V1. As discussed in Section \ref{5.2}, more low-quality crude in Cases 2 and 3 are considered, which leads to a small feasible region. Thus, repairing infeasible solutions becomes indispensable. As expected, Table \ref{result-3} demonstrates that DSEA/HR-V1 slightly improves performance compared to DSEA/HR-V2 through the assistance of the repair strategy. Similar observations are noted in the box plots in Fig. \ref{box-2}. The heuristic rules contribute to assisting DSEA/HR in the search for better solutions in case adequate computational resources are available. On the other hand, the repair strategy improves the computing performance of DSEA/HR in cases with smaller feasible regions. To gain insight into how these two components respectively work well in addressing the LSCOSPs, a detailed analysis of scheduling results is presented in the following section.

\subsubsection{\textbf{Effectiveness of the proposed heuristic rules in global search}} \label{eff-1}

\begin{figure*}[!t]
	\centering
	\includegraphics[scale= .59]{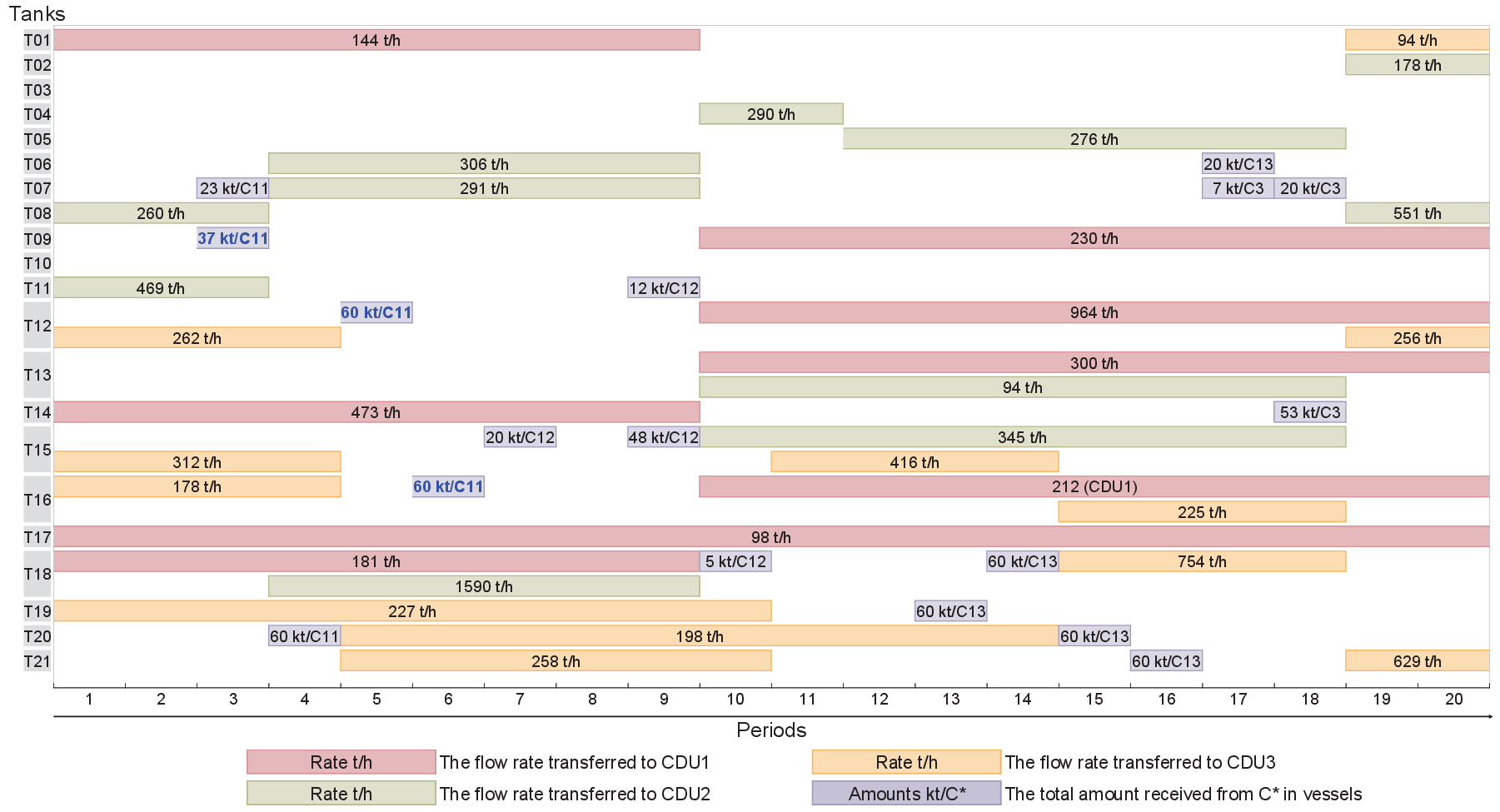}
	\caption{Gantt chart of Case 2 obtained by DSEA/HR. The blue numbers indicate instances of reasonable unloading operations.}\label{DSEA/HR}	
\end{figure*}

As mentioned earlier, the proposed DSEA/HR comprises two heuristic rules: one for assigning discharging tanks (denoted as Rule \uppercase\expandafter{\romannumeral1}) and the other for selecting charging tanks for each CDU (denoted as Rule \uppercase\expandafter{\romannumeral2}). To facilitate a better understanding and explanation of their effectiveness, we present a feasible schedule solved by the proposed DSEA/HR for Case 2 in Fig. \ref{DSEA/HR}. The horizontal axis represents the scheduling time, while the vertical axis indicates crude oil tanks.

\begin{figure*}[!t]
	\centering
	\includegraphics[scale= .59]{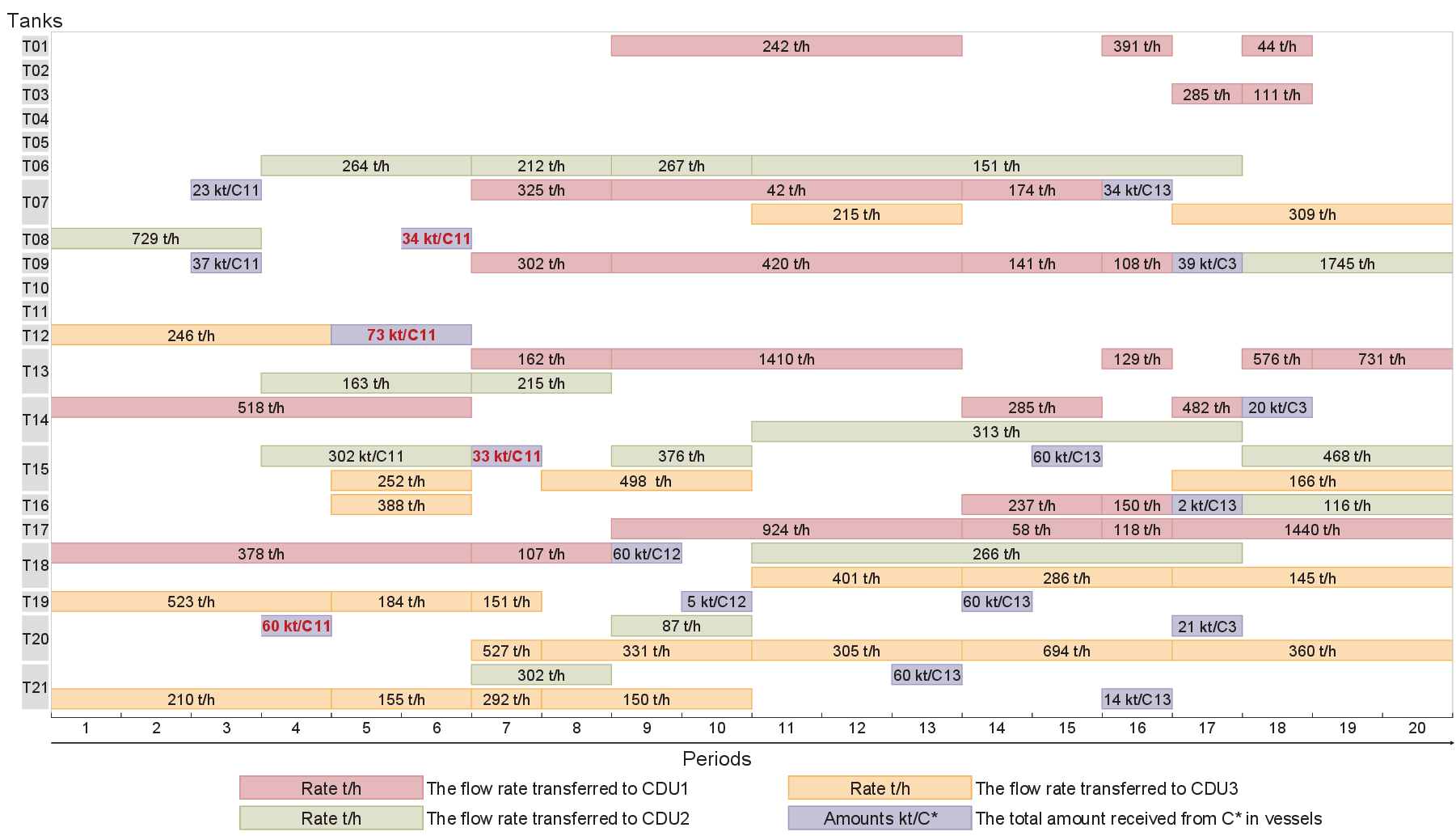}
	\caption{Gantt chart of Case 2 obtained by DSEA/HR-V1. The red numbers indicate instances of unreasonable unloading operations.}\label{DSEA/HR-V1}
\end{figure*}

\begin{figure*}[!h]
	\centering
	\includegraphics[scale= .59]{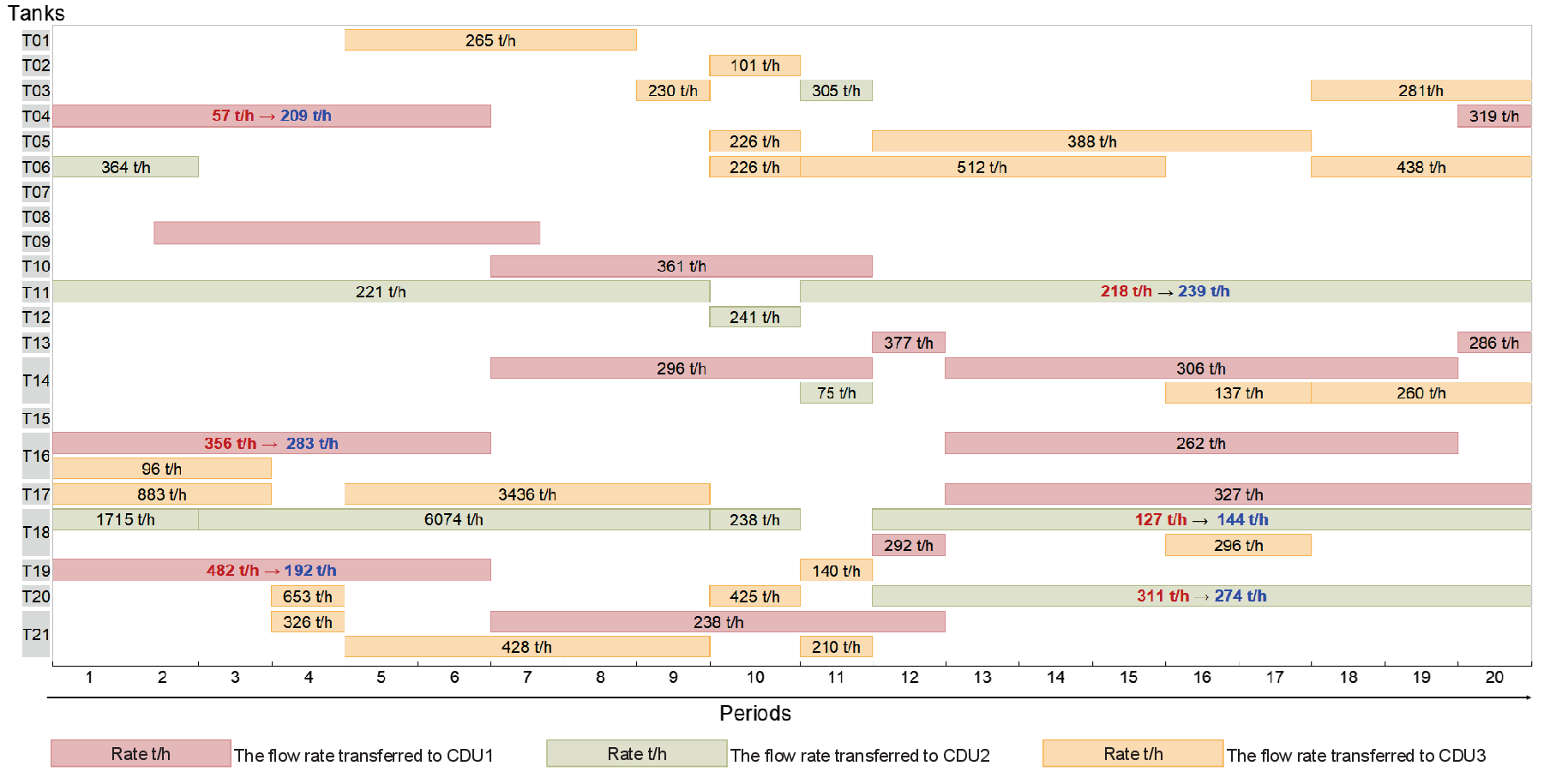}
	\caption{Gantt chart of Case 3 obtained by DSEA/HR and DSEA/HR-V2. The red numbers indicate the results obtained by DSEA/HR-V2, while the blue numbers indicate the results obtained by DSEA/HR.}\label{DSEA/HR-V2}
\end{figure*}

Regarding the effectiveness of Rule \uppercase\expandafter{\romannumeral1} on unloading operations, it can be observed from Fig. \ref{DSEA/HR} that the tanks (i.e., T09, T12, and T16) continuously charge to CDU1 between periods 10 and 20 after receiving crude C11 (indicated by blue numbers in Fig. \ref{DSEA/HR}). In contrast, Fig. \ref{DSEA/HR-V1}, obtained by DSEA/HR-V1, depicts a schedule that violates the property constraint due to the absence of heuristic rules for unloading operations. Specifically, the feed densities of CDU1 during periods 16-20 (i.e., 0.889, 0.9, 0.907, 0.909, 0.909 $g/cm^3$) exceed its upper limit (0.885 $g/cm^3$). This occurs because the low-density crude (i.e., C11) that satisfies the processing requirement is unloaded into tanks already containing other incompatible crudes (i.e., C2, C4, and C9) for CDU1. Apparently, the random unloading operations without the heuristic guidance contribute significantly to infeasibility and increased changeovers.

Furthermore, the performance of heuristic rules is also evident in charging operations. Fig. \ref{DSEA/HR-V1} shows an infeasible schedule solved by DSEA/HR-V1. A comparison between Fig. \ref{DSEA/HR} and Fig. \ref{DSEA/HR-V1} reveals that DSEA/HR achieves fewer changeovers than DSEA/HR-V1. This improved performance can be attributed to the heuristic rule designed for charging operations, which enhances the diversity of the tanks charging to a CDU. In contrast, due to the absence of this rule, several tanks (i.e., T14, T18, and T21) often charge a large amount of crude to two CDUs simultaneously. Although it is allowed in this problem, the drawback is that rapid consumption of inventory results in frequent changeovers between tanks and CDUs.

Based on the above comparisons, it can be concluded the proposed two heuristic rules are conducive to the performance improvement of the proposed DSEA/HR. The mechanism of heuristic rules, derived from empirical knowledge, can also be extended to more real-world instances.

\subsubsection{\textbf{Effectiveness of the repair strategy in local refinement}} \label{eff-2}

We have implemented a repair strategy during the local refinement stage to enhance the algorithm's search for feasible solutions. To elaborate on this, we have used Case 3 as an example, which poses a significant challenge due to the involvement of a large amount of low-quality crude types.

Our repair strategy, focusing on modifying the flow rate charging to CDUs, effectively solves this issue. Fig. \ref{DSEA/HR-V2} demonstrates this with a schedule involving all decisions of charging tanks, which is obtained by DSEA/HR-V2 and then modified by the repair strategy.
From the original results without repair in Fig. \ref{DSEA/HR-V2}, we can see that the sulfur content of the feed charging to CDU1 exceeds the upper limit during periods 1-6. In addition, the yield of naphtha produced by CDU2 falls below the lower bound during periods 12-20. These violations, however, can be completely eliminated through the local adjustment of the flow rate. Specifically, with the help of the repair strategy in the local refinement stage, the flow rate tied to decisions causing violations is further optimized in the continuous decision variable space. Following the flow rate optimization, we can observe from Fig. \ref{DSEA/HR-V2} that this schedule can be modified into a feasible solution that satisfies all constraints. As a result, implementing the repair strategy in the local refinement stage enhances DSEA/HR's capability to search for feasible regions, particularly in instances that involve multiple low-quality crudes.

Based on the above discussions, it can be concluded that both the heuristic rules and the repair strategy improve the optimization ability of the proposed DSEA/HR in discrete and continuous variable spaces, respectively. The integration of these strategies further enhances the solving capability of the EA-based algorithm, enabling it to find feasible solutions within an acceptable time.

\section{Conclusions} \label{section 5}
It is challenging to find the feasible solution of the large-scale crude oil scheduling problem, especially in a reasonable time. This paper investigated the LSCOSPs from a large marine-access refinery in practice. Firstly, the LSCOSPs accommodating the practical operating features were modeled. Following the model, we developed a novel EA-based framework called DSEA/HR to address the LSCOSPs.
Based on the empirical knowledge, we designed two heuristic rules for population initialization. One heuristic rule guides the decisions of unloading operations, while the other drives the decisions of charging operations. Moreover, to improve the efficacy of the algorithm in solving LSCOSPs with a small feasible region, we proposed a repair strategy to intensify the search for the continuous decision space.

In the experiments, we have compared the proposed algorithm with the state-of-the-art MINLP methods and metaheuristic algorithms on a number of LSCOSP instances. The comparative experimental results have demonstrated that DSEA/HR is superior to the competitors and is able to converge to the feasible region for all instances in an acceptable time. Meanwhile, the efficacy of the special designs is also further verified, including heuristic rules and the repair strategy. To the best of our knowledge, it is the first attempt to solve such large-scale crude oil scheduling problems that are very close to the operational features of an actual refinery. The applicability of the proposed method has also been verified in three practical cases.

A few issues remain to be further investigated in future work. First, while problem-specific knowledge is conducive to improving the efficacy of the algorithm, it cannot be directly applied to other scheduling problems. Second, the experimental results indicate that the algorithm does not guarantee the discovery of a feasible solution in each run. The feasible rate of algorithms tends to decrease as the complexity of the cases increases. To address these concerns, future research will focus on automatically extracting problem-specific knowledge by integrating emerging knowledge transfer techniques and incorporating efficient search strategies to enhance the stability of scheduling approaches.

\bibliographystyle{IEEEtran}
\bibliography{IEEEabrv,article-refs}

\end{document}